\documentclass{article}

\usepackage[final]{eiml_style_2025}

\usepackage{hyperref}
\usepackage{url}
\usepackage{graphicx}
\usepackage{tabularx} % Add this line to your document preamble
\usepackage{booktabs}
\usepackage[table]{xcolor} % put this in the preamble
\usepackage{amsmath}
\usepackage{amsmath}   % for \boldsymbol, align, equation etc.
\usepackage{amssymb}   % for \mathbb and many math symbols

\title{Credal Graph Neural Networks}

\author{Matteo Tolloso \& Davide Bacciu \\
Department of Computer Science \\
University of Pisa \\
Pisa, Italy \\
\texttt{matteo.tolloso@phd.unipi.it, davide.bacciu@unipi.it}
}

\begin{document}

\maketitle

\begin{abstract}
Uncertainty quantification is essential for deploying reliable Graph Neural Networks (GNNs), where existing approaches primarily rely on Bayesian inference or ensembles. In this paper, we introduce the first credal graph neural networks (CGNNs), which extend credal learning to the graph domain by training GNNs to output set-valued predictions in the form of credal sets. To account for the distinctive nature of message passing in GNNs, we develop a complementary approach to credal learning that leverages different aspects of layer-wise information propagation. We assess our approach on uncertainty quantification in node classification under out-of-distribution conditions. Our analysis highlights the critical role of the graph homophily assumption in shaping the effectiveness of uncertainty estimates. Extensive experiments demonstrate that CGNNs deliver more reliable representations of epistemic uncertainty and achieve state-of-the-art performance under distributional shift on heterophilic graphs.
\end{abstract}

\section{Introduction}

In safety-critical applications, a machine learning (ML) model must not only be accurate but also aware of its own limitations. The central challenge in making ML models reliable for real-world deployment is particularly acute for Graph Neural Networks (GNNs)~\citep{bacciu2020gentle}, where the inherent dependencies between data points violate the standard i.i.d. assumption. In a classical ML model, its confidence is often conflated into a single predictive probability, which fails to distinguish different sources of uncertainty \citep{gawlikowski2023survey}. This lack of specificity 
is critical on graphs, where a model can be uncertain about a node's class either because its local neighborhood is genuinely ambiguous (a property of the data) or because the node and its connections are entirely novel and outside the training distribution (a property of the model's knowledge) \citep{kendall2017uncertainties,hullermeier2021aleatoric}. To build more robust systems, we must formally disentangle the two primary sources of uncertainty: \emph{aleatoric uncertainty}, the inherent and often irreducible randomness in the data-generating process, and \emph{epistemic uncertainty}, which stems from the model's limited knowledge and is potentially reducible with additional data or improved models \citep{gal2016dropout, valdenegro2022deeper}. This decomposition underpins many modern approaches for uncertainty-aware learning and OOD detection \citep{kendall2017uncertainties}.

Despite the progress of existing approaches (see Appendix~\ref{sec:related}), a critical gap remains. Most current frameworks, particularly those based on message passing or uncertainty diffusion, implicitly assume graph homophily, i.e., that connected nodes tend to share similar features or labels \citep{ma_revisiting_2024}. As a result, their performance and the reliability of their uncertainty estimates deteriorate significantly in heterophilic settings, a challenge only recently beginning to be systematically addressed \citep{fuchsgruber2025uncertainty}. This highlights the need for new, robust uncertainty quantification frameworks that do not rely on homophily and can operate effectively across diverse graph structures.

\paragraph{Main Contributions.} We address the problem of uncertainty estimation in graph learning task, by  proposing an approach based on credal learning, and assessing specifically the underlooked aspect of graph heterophily. Our main contributions are: (i) we introduce \emph{Credal Graph Neural Networks (CGNNs)}, the first framework that extends credal learning to the graph domain; (ii) we develop a credal graph learning architecture that leverages layer-wise message passing to provide more faithful uncertainty representations; and (iii) we conduct extensive experiments on both homophilic and heterophilic benchmarks, showing that CGNNs achieve state-of-the-art performance in out-of-distribution detection and highlight the critical role of graph structure in shaping uncertainty estimates. The repository, which includes the implementation of our proposed Credal GNN and all other benchmarked models, can be accessed at the following anonymous link: \url{https://anonymous.4open.science/r/CGNN-EIML25}.

\section{Credal Graph Neural Networks}
\label{sec:methods}

This section introduces our Credal Graph Neural Network (CGNN) framework. We begin by outlining the principles of credal learning, which models uncertainty using sets of probability distributions to disentangle its aleatoric and epistemic sources. We then present our novel architecture for Credal Graph Learning by discussing on how the credal layer can be applied to graphs. 

\begin{figure}[h]
\centering
\includegraphics[width=\textwidth]{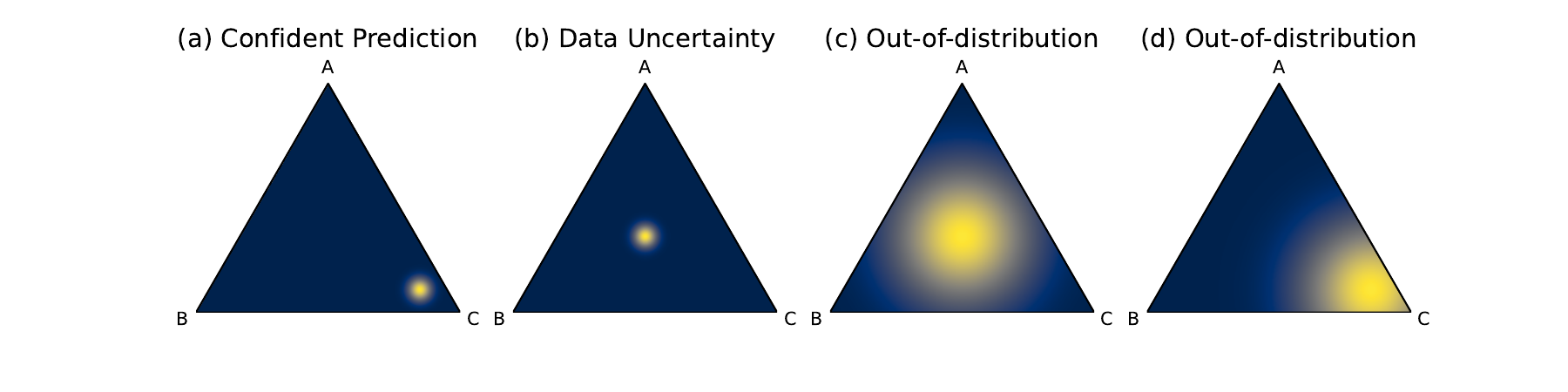}\vspace{-5mm}
\includegraphics[width=\textwidth]{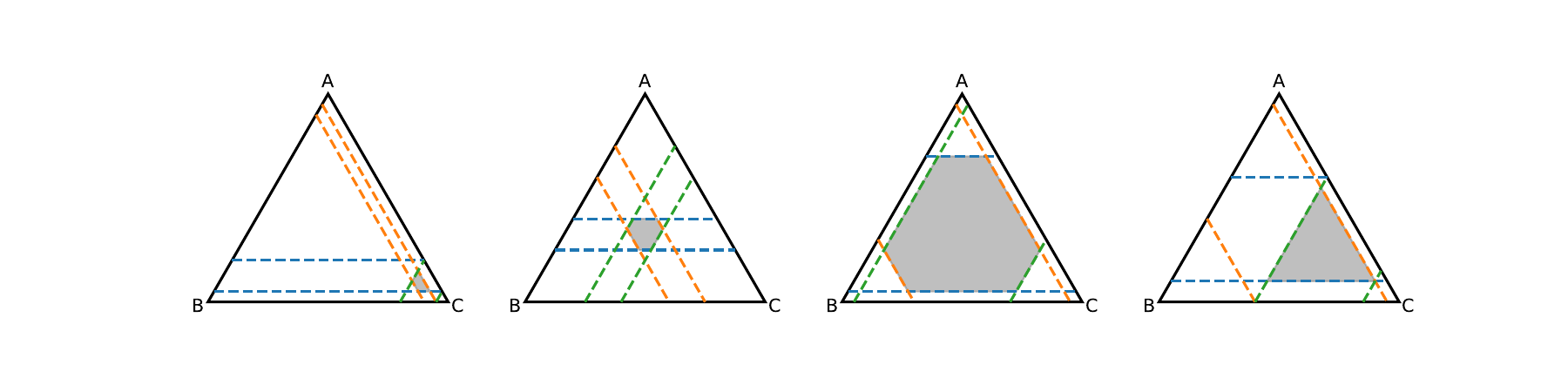}
\caption{Visualization of aleatoric uncertainty (AU) and epistemic uncertainty (EU) for a 3-class classification problem. The top row shows a Bayesian continuous representation, while the bottom row shows the corresponding credal set representation. 
(a) Low AU and low EU: the model is confident in its prediction. 
(b) High AU, low EU: the model attributes uncertainty to noisy data. 
(c) High AU and high EU: the model faces out-of-distribution data. 
(d) Low AU, high EU: the model encounters novel data and attributes uncertainty to its parameters rather than to noise. 
In the credal set representation (bottom row), the shaded regions inside the simplex correspond to lower and upper probability bounds for each class.}
\label{fig:2order_distributions}
\end{figure}

\paragraph{Credal Learning.} 
\label{sec:credal}
We begin by providing the basics of definitions behind our credal-based approach to uncertainty in graph learning. Credal learning has been proposed to address the limitations of BNNs and Ensemble-based approaches by explicitly modeling uncertainty as a convex set of probability distributions, known as a credal set $\mathcal{P}$ \citep{caprio2024credal}. In practice, the model constructs this set by outputting a probability interval for each class, with each interval being defined by a lower and an upper bound.
Figure~\ref{fig:2order_distributions} provides a visual example for 3-class classification task.

The shaded area represents the credal set, which can be interpreted as the region of the hypothesis that is coherent with the training data.

 Instead of averaging distributions, as needed by Bayesian Model Averaging or Ensembles (details in Appendix~\ref{sec:classical_uncertainty}), uncertainty can be defined by the size and location of $\mathcal{P}$ inside the simplex. Similarly to \citet{wang2024credalwrapper}, we quantify this using generalized entropy, which defines the total uncertainty TU and the aleatoric uncertainty AU as the maximum and minimum possible Shannon entropy for any distribution $p$ within the credal set \citep{abellan2006disaggregated}:
\begin{equation}
    \text{TU} := \overline{H}(\mathcal{P}) = \max_{\boldsymbol{p} \in \mathcal{P}} H(\boldsymbol{p}), \quad \text{AU} := \underline{H}(\mathcal{P}) = \min_{\boldsymbol{p} \in \mathcal{P}} H(\boldsymbol{p}). 
\end{equation}
The epistemic uncertainty is the difference, EU $:= \overline{H}(\mathcal{P}) - \underline{H}(\mathcal{P})$, which intuitively depends on the volume of the credal set and on its position inside the simplex. A confident prediction (Figure \ref{fig:2order_distributions}(a)) corresponds to a small set near a vertex. High data uncertainty (Figure \ref{fig:2order_distributions}(b)) is a small set near the center. High epistemic uncertainty is a large set covering a significant portion of the simplex (Figure \ref{fig:2order_distributions} (c and d)), indicating that many different probability distributions are considered plausible.

\paragraph{Credal graph layer.} 
\label{sec:credal_prediction}
To move beyond single-point probability predictions, we build on the {Credal Layer} proposed by \citet{wang2024credaldeep} as a replacement for the classification layer of a GNN. Let $\boldsymbol{z_v}$ be a generic embedding of node $v$ (as computed by some layer of a GNN), the Credal Layer would then take $\boldsymbol{z_v}$ as input and, for each of the $C$ classes, output two values: an interval midpoint $m^c_v$ and a half-length $h^c_v \geq 0$. Collecting these for all classes gives two vectors $\boldsymbol{m_v} = (m^1_v, \dots, m^C_v)$ and $\boldsymbol{h_v} = (h^1_v, \dots, h^C_v)$, each of length $C$. This is achieved using the transformation:

\begin{equation}
    \boldsymbol{m_v} = g({W} \times \boldsymbol{z_v} + \boldsymbol{b}), \quad \boldsymbol{h_v} = g'({W'} \times \boldsymbol{z_v} + \boldsymbol{b'}),
\end{equation}

where $g$ and $g'$ are activation functions, with $g' \geq 0$. These values define an interval for each class, $[\boldsymbol{a^L_v}, \boldsymbol{a^U_v}] := [\boldsymbol{m_v}-\boldsymbol{h_v}, \boldsymbol{m_v}+\boldsymbol{h_v}]$. To transform these into valid lower and upper probability bounds, $[\boldsymbol{q^L_v}, \boldsymbol{q^U_v}]$, that satisfy the necessary convexity conditions, a specialized Interval SoftMax activation \citep{wang2025creinns} is applied:

\begin{equation}
    q^{L_i}_v = \frac{\exp(a^{L_i}_v)}{\exp(a^{L_i}_v) + \sum_{k \neq i} \exp(a^{U_k}_v)}, \quad q^{U_i}_v = \frac{\exp(a^{U_i}_v)}{\exp(a^{U_i}_v) + \sum_{k \neq i} \exp(a^{L_k}_v)}.
\end{equation}
The resulting probability intervals define a credal set $\mathcal{P}_v = \{ \mathbf{q} | q^i \in [q^{L_i}_v, q^{U_i}_v], \sum_{i=1}^C q^i = 1 \}$, which constitutes the final credal prediction for node $v$. 

The training procedures, based on a Distributionally Robust Optimization (DRO) objective to ensure robustness to distributional shifts, is described in Appendix~\ref{sec:training}.

\paragraph{Information propagation on graphs.} A fundamental question is which representation $\boldsymbol{z_v}$ to use for a node $v$. For models processing independent and identically distributed (i.i.d.) data, the layer-wise processing can be described by the Data Processing Inequality (DPI) \citep{polyanskiy2017strong}. This principle states that the mutual information between a layer's latent representation $Z^{l}$ and the target label $Y$ cannot increase with network depth: $I(Y; Z^{l+1}) \le I(Y; Z^{l})$. Information is progressively filtered and compressed, and any information about $Y$ lost at layer $i$ cannot be recovered later.

The mechanism of information propagation within GNNs differs fundamentally from that in standard feed-forward networks. The DPI model of information flow is insufficient for GNNs, where the message-passing mechanism violates the i.i.d. assumption by design. A node's $v$ representation at layer $l$, namely ${z}_v^{l}$, is updated using information from its neighbors, effectively expanding its receptive field with each layer according to:
\begin{equation}
	{z}_v^{l} = \Phi^{l} \left( {z}_v^{l-1}, \Psi \left( \left\{ \Omega^{l}({z}_q^{l-1}) \mid q \in {N}_v\right\} \right) \right).
	\label{eq:gnn}
\end{equation}
In this formula, $N_v$ is the set of neighbors of node $v$, their representation in the layer $l-1$ is projected into a new space via the function $\Omega$ and aggregated into a single neighborhood representation using a permutation-invariant function $\Psi$ (such as max, mean, or sum). Finally, the function $\Phi$ combines this aggregated neighborhood representation with the embedding of node $v$ to produce the updated representation. 

\citet{fuchsgruber2025uncertainty} formalize this with a \textit{Data Processing Equality for Message Passing Neural Networks}, which shows that the information about a node's label $Y$ in its representation $Z^{l+1}$ at layer $l+1$ decomposes as:
\begin{equation}
    I(Y; Z^{l+1}) = I(Y; Z^{l}) - \Delta_{-}^{[{0:l}]} + \Delta_{+}^{l+1}.
\end{equation}
Here $I(Y; Z^{l})$ is the information about the target $Y$ contained in the node's own representation at the previous layer $l$. $\Delta_{-}^{[{0:l}]}$ is the \textit{relative information loss or recovery term}. It quantifies how much information about $Y$ from the node's ego-graph up to $l$ hops is lost (while also accounting for the recovery of previously lost information retained in node’s neighbors) during the $l+1$-th message-passing step. $\Delta_{+}^{l+1}$ is the \textit{information gain term}, representing new, additional information about $Y$ that is incorporated from the newly included $l+1$-hop neighbors.

This decomposition is key to understanding the difficulty of graph GNNs in particular in heterophilic learning, as observed by many authors \citep{micheli2023addressing, liang2023predicting, ma2021homophily, zhu2020beyond, luan2024heterophilic, lim2021large}. In homophilic settings, neighbors are semantically similar, so the information gain $\Delta_{+}^{l+1}$ is often low (i.e. information tends to be redundant). In contrast, in heterophilic settings, neighbors are semantically \textit{different}, meaning the information gain $\Delta_{+}^{l+1}$ can be substantial at each layer. Each latent representation $\boldsymbol{z_v^{l}}$ thus provides unique and crucial information. The fundamental flaw of standard message-passing in this regime is that it acts as a local smoothing operator, indiscriminately mixing the central node's signal with these diverse neighboring signals. This corruption leads to representations of different classes becoming indistinguishable.

Motivated by these observations we propose to use in input to the credal layer $\boldsymbol{z^{\text{joint}}_v}$, that is the joint latent representation of a node $v$, computed by concatenating the node's embeddings from all layers:
\begin{equation}
    \boldsymbol{z_v^{\text{joint}}} = [\boldsymbol{z_v^{0}} || \boldsymbol{z_v^{1}} || \dots || \boldsymbol{z_v^{L}}],
\end{equation}
where $\boldsymbol{z_v^{0}}$ are the initial (input) node features. 

\begin{figure}[t]
    \centering
    \includegraphics[width=.9\textwidth]{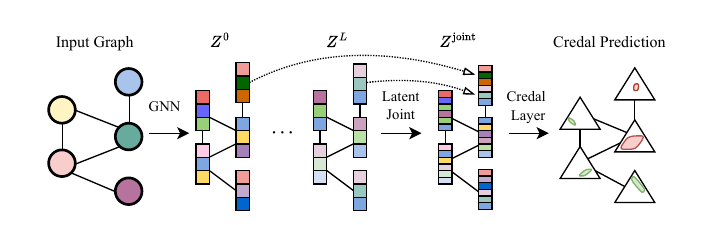}
    \caption{Overview of the proposed credal prediction framework. An input graph is processed by a Graph Neural Network (GNN), which iteratively aggregates node representations across layers ($Z^0, \ldots, Z^L$), gaining and losing information at each step. The latent joint representation $Z^{\text{joint}}$ allows recovery of the full trajectory of each node across layers. A credal layer then maps $Z^{\text{joint}}$ to a credal prediction, providing an informative representation of uncertainty. Green credal sets correspond to reliable predictions, while red credal sets indicate nodes that are likely out-of-distribution (OOD) due to high aleatoric or epistemic uncertainty.}
    \label{fig:full_architecture}
\end{figure}

By using $\boldsymbol{z^{\text{joint}}_v}$ as node's $v$ embedding, we ensure the credal prediction is conditioned on all the information the GNN has processed, allowing for a more faithful and robust estimation of both the prediction and its associated uncertainty. 
The full architecture, named \texttt{CredalLJ} for Latent Joint, is depicted in Figure \ref{fig:full_architecture}.

\paragraph{Ablated models.} To isolate the effect of each component of our system we conducted a series of ablation studies. \texttt{Credal Final} removes the latent joint representation forcing the credal layers to rely on the standard final embedding representation. \texttt{Credal Ensemble} is a post-hoc credal methods that do not require specialized training procedure nor specialized credal layer. Finally \texttt{KNNLJ} isolates the effects of latent joint representation with a non-credal method. The ablated models are described in details in Appendix~\ref{sec:ablation_study}.

\section{Experimental Analysis}

In this section, we empirically validate our proposed Credal Graph Neural Networks. We define the experimental setting for out-of-distribution (OOD) node detection and we present the results. The datasets used for our evaluation are described in details in Appendix~\ref{sec:dataset}. The baselines for comparison are detailed in Appendix~\ref{sec:baselines}, while the ablated models in Appendix~\ref{sec:ablation_study}.

\paragraph{Setting.} 
\label{sec:settings}

We address the problem of node-level out-of-distribution (OOD) detection in a transductive setting using a \emph{Leave-Out-Class} strategy during training. Let $G = (\mathcal{V}, \mathcal{E}, X)$ denote a single attributed graph, where $\mathcal{V}$ is the set of nodes, $\mathcal{E}$ is the set of edges, and $X \in \mathbb{R}^{|\mathcal{V}| \times d}$ is the node feature matrix. The set of all node classes is partitioned into in-distribution (ID) classes $\mathcal{C}_{ID}$ and out-of-distribution (OOD) classes $\mathcal{C}_{OOD}$, such that $\mathcal{C}_{ID} \cap \mathcal{C}_{OOD} = \emptyset$. 

The node set $\mathcal{V}$ is divided into training, validation, and test subsets. During training, nodes belonging to OOD classes in $\mathcal{C}_{OOD}$ are masked, i.e., their labels are treated as unknown. Consequently, the model is trained to solve a classification problem over only the ID classes $\mathcal{C}_{ID}$. Importantly, in this setting, the true labels of test nodes may include both ID and OOD classes, i.e., the test classes form a superset of the training classes. 

The objective is to learn a model that classify the in-distribution nodes into their respective classes within $\mathcal{C}_{ID}$ and produce a high uncertainty score for nodes belonging to the OOD classes.

\paragraph{Results.}
\label{sec:results}
The OOD detection performance of our proposed methods and all baselines is presented in Table~\ref{tab:ood_comparison}, with results reported in AUROC. The experiments span six diverse datasets, including both homophilic and highly heterophilic graphs, allowing for a comprehensive evaluation of robustness to different graph structures. Additional results are reported in Appendix~\ref{sec:additional_results}.

\definecolor{lightgray}{gray}{0.9}
\begin{table}[h]
\caption{OOD detection performance across all datasets (AUROC $\uparrow$). For models that allow uncertainty disentanglement, both aleatoric and epistemic uncertainty scores are reported (aleatoric/epistemic). Best results are highlighted in \textbf{bold}, second-best results are \underline{underlined}, and third-best results are shown with a light gray background. Results with variance are reported in Appendix \ref{sec:additional_results}.}
\label{tab:ood_comparison}
\centering % Center the resizebox
\resizebox{\textwidth}{!}{%
\begin{tabular}{lcccc|cc}
\toprule
Method & Chameleon & Squirrel & ArXiv & Patents & Coauthor & Reddit2 \\
\midrule

Energy & 58.25 & 44.47 & 50.16 & 43.33 & \colorbox{lightgray}{94.47} & 43.93 \\
KNN & 56.81 & 52.79 & 56.86 & 52.55 & 88.23 & 65.63 \\
ODIN & 57.98 & 47.28 & 48.16 & 43.21 & 94.17 & 43.09 \\
Mahalanobis & 51.82 & 53.79 & 59.52 & 58.72 & 82.49 & \colorbox{lightgray}{68.98} \\
GNNSafe & 50.42 & 35.88 & 35.30 & 27.35 & \underline{94.82} & 61.99 \\
Classical ensemble & 74.00/30.22 & 58.32/59.13 & 58.20/\colorbox{lightgray}{65.45} & 48.23/60.35 & \textbf{95.30}/94.81 & 58.84/\textbf{72.09} \\

KNNLJ & 70.06 & \colorbox{lightgray}{71.06} & 45.35 & 46.74 & 32.37 & 36.55 \\
\midrule
Credal final & \underline{76.29}/67.27 & \underline{75.02}/65.85 & 64.65/\underline{65.66} & 68.92/\underline{69.73} & 74.80/50.74 & \underline{70.57}/69.35 \\
Credal ensemble & \colorbox{lightgray}{74.98}/29.54 & 59.03/53.66 & 58.05/50.97 & 47.41/\colorbox{lightgray}{64.64} & 93.85/93.72 & 61.16/57.72 \\
CredalLJ & 72.37/\textbf{77.67} & 73.89/\textbf{77.04} & \textbf{65.77}/63.79 & \textbf{70.78}/60.06 & 86.58/70.88 & 66.35/67.31 \\
\bottomrule
\end{tabular}%
}
\end{table}

The results highlight the limitations of existing uncertainty quantification methods, particularly on heterophilic graphs. Our \texttt{CredalLJ} achieves the best performance on all heterophilic benchmarks (Chameleon, Squirrel, ArXiv, Patents), while remaining competitive on homophilic ones, being only a few percentage points below the \texttt{Classical Ensemble}, which stands out as the best baseline.

The \texttt{Credal Final} consistently ranks second on all heterophilic benchmarks, being only 1--2 percentage points below the \texttt{CredalLJ}, which highlights the importance of the Credal module.

\texttt{Credal Ensemble} and \texttt{KNNLJ} performed worse than expected. In particular, \texttt{KNNLJ} proved to be an unreliable uncertainty estimator, achieving good performance only on two datasets, while \texttt{Credal Ensemble} slightly underperformed the \texttt{Classical Ensemble}, with the exception of the Patents dataset. This suggests that a simple post-hoc application of credal theory offers limited gains without architectural and training modifications.

As expected, the \texttt{Classical Ensemble} outperformed all baselines on both homophilic and heterophilic datasets. However, the baselines maintained competitiveness in the homophilic setting.

On the homophilic {Coauthors} dataset, many standard post-hoc methods, including \texttt{Energy}, and the graph-specific \texttt{GNNSafe}, achieve strong performance. Similarly for Reddit2 where \texttt{Mahalanobis} and \texttt{KNN} are competitive. However, the performance of nearly all single-model baselines collapses dramatically on the heterophilic datasets, especially for \texttt{GNNSafe} which relies on homophilic smoothing kernels. 

A deeper analysis of the results reveals several intriguing patterns. 
First, the optimal type of uncertainty for OOD detection is highly dataset-dependent. 
For our state-of-the-art \texttt{CredalLJ} model, {epistemic uncertainty} provides the strongest signal on the smaller heterophilic graphs, whereas {aleatoric uncertainty} is more effective on the larger ones. 
This suggests that as the scale of the graph increases, the most reliable OOD indicator may shift from the model's own ignorance (epistemic) to the inherent ambiguity of the data itself (aleatoric).

We observe that, in general, models rely on different uncertainty types for different graphs. This highlights the practical value of our proposed methods, which can successfully disentangle and leverage both aleatoric and epistemic uncertainty to adapt to a wide variety of graph properties and complexities. This interplay between aleatoric and epistemic uncertainty in the context of OOD detection has been noted in prior work \citep{stadler2021graph, fuchsgruber2024energy, fuchsgruber2025uncertainty} and requires further study.

\section{Conclusions}

In this paper, we introduced Credal Graph Neural Networks, the first framework to extend credal learning to the graph domain for robust uncertainty quantification. 
Our approach addresses the limitations of existing methods, which often falter on graphs that violate the homophily assumption. 
We proposed a novel architecture that leverages a joint latent representation of the full information propagation trajectory across all GNN layers. 
This, combined with a training objective inspired by Distributionally Robust Optimization, allows our models to produce set-valued predictions in the form of credal sets, enabling a principled disentanglement of aleatoric and epistemic uncertainty.

Our experiments demonstrate the effectiveness of our approach, particularly on challenging heterophilic graphs. 
The proposed framework sets a new state-of-the-art for out-of-distribution detection across all heterophilic benchmarks, where the performance of most single-model baselines collapses dramatically. 
Furthermore, our models remain highly competitive on homophilic datasets, showcasing their robustness and versatility across diverse graph structures. 
These results highlight the importance of both the novel joint latent architecture and the specialized credal training procedure for achieving reliable uncertainty estimates.

By introducing the first credal GNNs, this work opens a new and promising research direction for uncertainty-aware graph machine learning. 
Our findings also underscore the need for a deeper investigation into the complex interplay between aleatoric and epistemic uncertainty in the context of OOD detection. 
Future work should therefore focus on better understanding this relationship, exploring how these distinct sources of uncertainty can be optimally combined or selectively leveraged to build more reliable and trustworthy GNNs for safety-critical applications.

\paragraph{Reproducibility statement}

To ensure the reproducibility of our research, we have made our complete source code and experimental setup publicly available. The repository, which includes the implementation of our proposed Credal GNN and all other benchmarked models, can be accessed at the following anonymous link: \url{https://anonymous.4open.science/r/CGNN-EIML25}.
The repository contains detailed instructions for setting up the required environment and scripts to replicate the experiments presented in this paper.
Furthermore, all datasets utilized in our study are established, open-source benchmarks from the graph machine learning community. These datasets can be retrieved from their original publications, which are cited in our manuscript where each dataset is first introduced. We believe that the provided code and the public nature of the datasets offer sufficient resources for the research community to verify our findings and build upon our work.

\bibliographystyle{plainnat}
\bibliography{bib}

\appendix

\newpage
\section{LLM Usage}
\label{app:llm_usage}

Large Language Models (LLMs) were used as general-purpose assistive tools during the preparation of this paper. Their role was limited to two aspects:  
(i) improving the writing quality (e.g., grammar correction, rephrasing for clarity, and suggesting minor improvements to text structure), and  
(ii) assisting with retrieval and discovery of related work (e.g., suggesting potentially relevant references for further manual inspection by the authors).  

All content generated or suggested by LLMs was carefully reviewed, verified, and, where necessary, edited by the authors to ensure accuracy and faithfulness to the intended scientific contributions. LLMs were not used to generate scientific content, including research ideas, hypothesis formulation, experimental design, data analysis, or interpretation of results.  

The authors take full responsibility for all parts of the paper.

\section{Related Works} 
\label{sec:related}
Quantifying uncertainty in Graph Neural Networks (GNNs) \citep{bacciu2020gentle} is a relevant and rapidly developing research direction, with several methods being proposed recently to capture the reliability of GNN predictions \citep{wang2024uncertainty, chen2024uncertainty}. The inherent dependencies among nodes in a graph introduce unique challenges, leading to a diverse landscape of quantification strategies. Existing approaches can be broadly organized into three main families, each with distinct computational and modeling characteristics.

The most computationally efficient family consists of \textit{Single Deterministic Models}. The simplest methods in this category are post-hoc heuristics applied to a standard GNN's output, such as using the maximum softmax probability as a measure of confidence or calculating the predictive entropy. While easy to implement, these approaches provide a single, undifferentiated measure of uncertainty and are often sensitive to model miscalibration \citep{guo2017calibration}. A more principled deterministic approach is evidential deep learning, where the GNN learns to output the parameters of a higher-order Dirichlet distribution. This allows for directly modeling uncertainty over the categorical output space in a single forward pass, providing a richer characterization of the predictive uncertainty \citep{zhao_uncertainty_2020, stadler_graph_2021}. 

A second major paradigm is \textit{Bayesian Graph Neural Networks (Bayesian GNNs)}, which adapt the Bayesian framework specifically to graph-structured data. In these models, one places priors not only over standard network parameters (e.g., weights) but also over graph-specific components such as edge connectivity or node feature propagation. For example, adaptive connection sampling \citep{hasanzadeh2020bayesian} treats edges (or adjacency masks) probabilistically, and inference techniques like Monte Carlo Dropout~\citep{gal2016dropout} or Variational Inference \citep{hoffman2013stochastic} are used to sample over both weights and possible graph instantiations during prediction. This both reflects uncertainty in what the graph structure (and features) imply, and in the model's parameters. Works such as \citet{munikoti2023general} illustrate how Bayesian methods in graphs model uncertainty over graph structure, node features, and edge sampling, not just traditional weights. This provides a theoretically grounded framework for capturing uncertainty stemming from the model's own parameters and from uncertainty in the graph data (structure or features), but at the cost of higher computational complexity \citep{jia2020efficient} and the challenge of defining suitable priors for graph structures.

 Bridging the gap between theoretical rigor and practical scalability, \textit{Ensemble Methods} have become a powerful and popular alternative. This approach averages the predictions from multiple GNNs that are trained independently. Uncertainty is then quantified by measuring the disagreement or variance among the predictions of the individual models in the ensemble \citep{mallick2022deep, busk2023graph}. While empirically powerful and often outperforming more complex Bayesian methods, ensembles also carry a significant computational and memory burden, as they require training and storing multiple full models.

\section{Model Training}
\label{sec:training}

To train a model capable of output predictions in the form of a credal set we frame our problem within the setting of learning from a collection of data sets $\{\mathcal{D}_1, \dots, \mathcal{D}_Q\}$, $\mathcal{D}_i = \{\boldsymbol{(x_{i,1}}, \boldsymbol{y_{i,1}}), \dots, (\boldsymbol{x_{i,n_i}}, \boldsymbol{y_{i,n_i}})\}$, where each $\mathcal{D}_i$ is issued from a different ``domain" characterized by its own unknown data-generating probability distribution \citep{caprio2024credal}. The idea is that by leveraging the available evidence $\mathcal{D}_i$ the model is able to elicit a credal set that contains the true data generating process for a new set of data $\mathcal{D}_{Q+1}$. In practice, however, we are often limited to a single training set $\mathcal{D}$, which can be viewed as an aggregation of data from these various underlying domains. We adopt, following \citet{wang2024credaldeep}, a strategy inspired by Distributionally Robust Optimization (DRO) \citep{ben2013robust}. Instead of minimizing only the classical empirical risk, the objective is to also find model parameters $\boldsymbol{\theta}$ that minimize the worst-case risk over an uncertainty set of distributions $\mathcal{U}$ around the empirical distribution $\hat{P}$:
\begin{equation}
\label{eq:dro_loss}
    \min_{\boldsymbol{\theta} \in \Theta} \left\{ \sup_{U \in \mathcal{U}} \mathbb{E}_{(\boldsymbol{x},\boldsymbol{y})\sim U}[\mathcal{L}(h_{\boldsymbol{\theta}}, (\boldsymbol{x},\boldsymbol{y}))] \right\},
\end{equation}
where $\mathcal{L}$ is a given loss function and $h_{\boldsymbol{\theta}}$ is the hypothesis parameterized by $\boldsymbol{\theta}$. 
Training the model to be robust against this internally-generated adverse distribution compels it to produce predictions in the form of a credal set, represented by a collection of probability intervals.

Following heuristics \citep{huang2022two, oren2019distributionally}, the DRO loss can be approximated as: 
\begin{equation}
\label{eq:loss}
\mathcal{L}\;=\; 
\frac{1}{N}\sum_{n=1}^N \text{CE}(\boldsymbol{q^{U}_n}, \boldsymbol{y_n})
\;+\;
\frac{1}{\delta N} \sum_{n \in \mathcal{H}_\delta} \text{CE}(\boldsymbol{q^{L}_n}, \boldsymbol{y_n}),
\end{equation}
where
\begin{equation}
\mathcal{H}_\delta = 
\operatorname*{arg\,top}_{\;\delta N}\Big\{ \text{CE}(\boldsymbol{q^{L}_n}, \boldsymbol{y_n}) \;\big|\; n = 1, \dots, N \Big\},
\end{equation}
$N$ is the number of training points, CE is the cross-entropy and $\delta$ is an hyper-parameter that indicates the proportion of difficult training examples to consider for an estimate of test distribution divergence. The first component of Equation \ref{eq:loss} relies on the training distribution as is, and therefore tends to encourage more \emph{optimistic} or upper-bound predictions for the class scores ($\boldsymbol{q^U}$). In contrast, the second component emphasizes the boundary cases and training outliers by operating on $\boldsymbol{q^L}$, simulating potential shifts between training and test distributions. This encourages the model to produce more \emph{pessimistic} or lower-bound predictions. As a consequence, the width between $\boldsymbol{q^U}$ and $\boldsymbol{q^L}$ reflects the model's ignorance about how much the future test distribution may differ from the training distribution. By leveraging the outlier cases observed at training time we approximate the DRO loss (Equation \ref{eq:dro_loss}), allowing the model to better estimate the uncertainty it may encounter at test time.

\section{Additional experimental details}

 \subsection{Datasets}
\label{sec:dataset}

We evaluate our method on a diverse suite of six benchmark graph datasets, encompassing various domains and structural properties, as summarized in Table \ref{tab:dataset_summary}.

\begin{table}[h]
\caption{Statistics and OOD splits for the benchmark datasets. Validation and test set sizes are reported as in-distribution (ID) / out-of-distribution (OOD).}
\label{tab:dataset_summary}
\centering
\resizebox{\textwidth}{!}{%
\begin{tabular}{lrrrc llrrr}
\toprule
Dataset & \# Nodes & \# Edges & \# Classes & Homophily & OOD Classes & ID Classes & Train (ID) & Val (ID/OOD) & Test (ID/OOD) \\
\midrule
Chameleon & 2,277 & 31,421 & 5 & He & \{0,1\} & \{2,3,4\} & 647 & 438 / 291 & 276 / 180 \\
Squirrel & 5,201 & 198,493 & 5 & He & \{0,1\} & \{2,3,4\} & 1,515 & 982 / 682 & 622 / 419 \\
ArXiv & 169,343 & 1,166,243 & 5 & He & \{0,1\} & \{2,3,4\} & 69,523 & 23,415 / 10,453 & 23,245 / 10,625 \\
Patents & 2,923,922 & 13,975,788 & 5 & He & \{0,1\} & \{2,3,4\} & 1,053,055 & 351,259 / 233,525 & 350,730 / 234,055 \\
Coauthor & 18,333 & 163,788 & 15 & Ho & \{0,\dots,3\} & \{4,\dots,14\} & 8,813 & 2,907 / 759 & 2,964 / 704 \\
Reddit2 & 232,965 & 23,213,838 & 41 & Ho & \{0,\dots,10\}& \{11,\dots,40\} & 109,517 & 17,343 / 6,356 & 40,601 / 14,733 \\
\bottomrule
\end{tabular}%
}
\end{table}

The \texttt{Coauthors} dataset is a computer science co-authorship network where nodes represent authors, connected by an edge if they have co-authored a paper \citep{shchur2018pitfalls}. Node features are derived from paper keywords, and the task is to predict each author's primary field of study. The \texttt{Chameleon} and \texttt{Squirrel} datasets are Wikipedia networks where nodes are web pages linked by hyperlinks \citep{rozemberczki2021multi}. Their features are derived from informative nouns on each page, and the classification task is to categorize pages based on their average monthly traffic. \texttt{Reddit2} is a dataset constructed from Reddit posts, where nodes represent individual posts with features generated from text embeddings \citep{hamilton2017inductive}. An edge connects two posts if the same user has commented on both, and the prediction task is to identify the subreddit to which each post belongs. \texttt{ArXiv} is a citation network where nodes are academic papers and directed edges represent citations \citep{hu2020open, ma_revisiting_2024}. Node features are embeddings of the paper's title and abstract, and the objective is to predict the publication year. Finally, \texttt{Patents} is a citation network of U.S. utility patents, where each node is a patent and edges indicate citations between them \citep{leskovec2005graphs, lim2021large, ma_revisiting_2024}. Features are derived from patent metadata, and the task is to predict the patent's grant date.

\subsection{Baselines}
\label{sec:baselines}
We evaluate our proposed credal learning approaches against a suite of established baselines for uncertainty estimation and OOD detection~\citep{ma_revisiting_2024}. These include a family of widely-used post-hoc methods that operate on a single pre-trained GNN, such as the \texttt{Energy}-based score, which is calculated from the pre-softmax logits \citep{liu2020energy}; \texttt{ODIN}, which applies temperature scaling and input perturbations \citep{liang2017enhancing}; and the \texttt{Mahalanobis} baseline, which measures the distance of a test sample's latent representation from the training data's class-conditional distributions \citep{lee2018simple}. Similarly, we employ a \texttt{K-Nearest Neighbors (KNN)} approach, where uncertainty is derived from the average latent-space distance to the nearest training samples \citep{sun2022out}. We also compare against a method specifically designed for graph data, namely \texttt{GNNSafe}, a graph-specific Energy-Based Model that incorporates a label propagation mechanism \citep{wu2023energy}.

However, the de-facto reference model against which new uncertainty quantification methods are measured is the \texttt{Classical Ensemble}. Despite its conceptual simplicity, this approach consistently achieves state-of-the-art performance across a wide range of tasks and datasets, making it the target method to outperform \citep{lakshminarayanan2017simple, gustafsson2020evaluating, ovadia2019can, abe2022deep}. Its strength, lies in its combination of high performance with practical simplicity: it is easy to implement, scales effectively, and is largely hyperparameter-free, requiring only the independent training of multiple standard models. 

Additional details on model selection are reported in Appendix \ref{sec:model_selection}.

\subsection{Ablated Models}
\label{sec:ablation_study}

To better understand the effect of each component in our proposed \texttt{CredalLJ} GNN model, we conduct a series of ablation studies.

To isolate the effect of the credal output layer in a more standard GNN setting, we evaluate a variant where it is applied only to the final hidden layer. In this model, which we term \texttt{Credal Final}, we remove the concatenation of embeddings from all layers. The credal output layer, takes only the final latent representation, $\boldsymbol{z_v^{L}}$, of the GNN backbone as its input. 
The \texttt{Credal Final} removes the effect of the joint trajectory of the embedding so that the credal module is constrained to rely solely on the final node representation.

To test whether the performance of our model stems primarily from its latent joint representation rather than the credal module itself, we conduct a second ablation. We replace our credal output layer with a simple KNN density estimator, a method inspired by the Joint Latent Density Estimation (JLDE) principle recently proposed by \citet{fuchsgruber2025uncertainty}. Specifically, we compute the OOD score based on the k-Nearest Neighbors (KNN) distance within the joint latent space, $\boldsymbol{z_v^{\text{joint}}}$, used by our full model. We refer to this model as \texttt{KNNLJ}.

The credal graph learning approach requires modifying the original GNN architecture, hence it cannot be applied post-hoc to already trained model. As an alternative method, we also explore a post-hoc approach
%As an alternative to modifying the network architecture, we also explore a post-hoc method 
for generating credal predictions from a set of GNNs trained for a standard classification task. Inspired by \citep{wang2024credalwrapper}, we define a \texttt{Credal Ensemble} leveraging an ensemble of vanilla GNNs that requires no changes to their training. For a given input node $v$, each model $h_m$ produces a standard probabilistic prediction in the form of a softmax vector. The core principle of the \texttt{Credal Ensemble} is to interpret the set of these predictions, $\{\mathbf{p_1}, \dots, \mathbf{p_M}\}$, as the vertices of a credal set. The final credal prediction for node $v$, $\mathcal{P}_v$, is therefore defined as the convex hull of the ensemble's outputs:
\begin{equation}
    \mathcal{P}_v = \text{Conv}(\{\mathbf{p_1}, \dots, \mathbf{p_M}\}).
\end{equation}
This resulting set contains all possible convex combinations of the individual model predictions, thereby capturing the epistemic uncertainty expressed through the disagreement among the ensemble members. This method offers a straightforward way to obtain a credal prediction from any existing pre-trained GNN ensemble. This \texttt{Credal Ensemble} removes the specialized training procedure for the credal module, aiming to estimate the credal set from a set of standard prediction.

\section{Additional results}
\label{sec:additional_results}

\begin{table}[h]
\caption{In-distribution (ID) test classification performance (F1-score $\uparrow$). 
For our credal models, we report results based on two predictions, using the vectors $\boldsymbol{q_L}$ and $\boldsymbol{q_U}$. 
We report mean $\pm$ standard deviation over 5 runs. 
Best results are highlighted in \textbf{bold}, second-best results are \underline{underlined}, and third-best results are shown with a light gray background.}
\label{tab:id_classification}
\centering
\resizebox{\textwidth}{!}{%
\begin{tabular}{llcccccc}
\toprule
& Method & Chameleon & Squirrel & ArXiv & Patents & Coauthor & Reddit2 \\
\midrule
& Vanilla GNN & \textbf{57.61 $\pm$ 0.98} & \textbf{46.46 $\pm$ 2.25} & \underline{51.81 $\pm$ 0.90} & \textbf{53.35 $\pm$ 0.43} & \textbf{95.23 $\pm$ 1.41} & \textbf{79.47 $\pm$ 0.36} \\
& Credal Final ($\boldsymbol{q_L}$) & \cellcolor{gray!15}{53.99 $\pm$ 1.42} & 39.55 $\pm$ 0.74 & 51.03 $\pm$ 0.40 & 51.20 $\pm$ 0.41 & \cellcolor{gray!15}{50.53 $\pm$ 0.89} & \underline{27.38 $\pm$ 0.36} \\
& Credal Final ($\boldsymbol{q_U}$) & \underline{55.43 $\pm$ 0.38} & 39.55 $\pm$ 0.43 & 51.03 $\pm$ 0.42 & \underline{51.94 $\pm$ 0.40} & 28.60 $\pm$ 0.81 & 24.63 $\pm$ 0.36 \\
& CredalLJ ($\boldsymbol{q_L}$) & 42.75 $\pm$ 0.41 & \underline{39.71 $\pm$ 0.45} & \textbf{52.39 $\pm$ 0.39} & 49.34 $\pm$ 0.40 & \underline{75.44 $\pm$ 0.98} & 23.83 $\pm$ 0.37 \\
& CredalLJ ($\boldsymbol{q_U}$) & 36.96 $\pm$ 0.39 & 36.17 $\pm$ 1.44 & \cellcolor{gray!15}{51.01 $\pm$ 0.40} & 49.34 $\pm$ 0.41 & 43.76 $\pm$ 0.42 & \cellcolor{gray!15}{27.29 $\pm$ 0.37} \\
\bottomrule
\end{tabular}%
}
\end{table}

We report the in-distribution (ID) classification performance, measured by the F1-score, in Table~\ref{tab:id_classification}. The results show that the {Vanilla GNN} baseline generally outperforms our credal models on ID data. This outcome is expected and highlights a known trade-off in robust machine learning: the Vanilla GNN is optimized solely for empirical risk minimization, maximizing performance on data from the training distribution. Our credal models, in contrast, are regularized via a Distributionally Robust Optimization (DRO) objective to enhance out-of-distribution (OOD) robustness, which naturally results in more conservative predictions and a slight decrease in ID performance.

\begin{table}[t]
\caption{OOD detection performance across all datasets (AUROC$\uparrow$). We report the mean and standard deviation over 5 runs. For models that allow uncertainty disentanglement, we report aleatoric (AU) and epistemic (EU) scores separately.}
\label{tab:ood_comparison_full}
\centering
\resizebox{\textwidth}{!}{%
\begin{tabular}{llcccccc}
\toprule
& Method & Chameleon & Squirrel & ArXiv & Patents & Coauthor & Reddit2 \\
\midrule
& Energy & 58.25 $\pm$ 4.63 & 44.47 $\pm$ 1.62 & 50.16 $\pm$ 4.71 & 43.33 $\pm$ 3.48 & \colorbox{lightgray}{94.47 $\pm$ 0.45} & 43.93 $\pm$ 1.24 \\
& KNN & 56.81 $\pm$ 6.28 & 52.79 $\pm$ 1.55 & 56.86 $\pm$ 1.39 & 52.55 $\pm$ 1.24 & 88.23 $\pm$ 2.39 & 65.63 $\pm$ 1.50 \\
& ODIN & 57.98 $\pm$ 2.46 & 47.28 $\pm$ 2.05 & 48.16 $\pm$ 2.37 & 43.21 $\pm$ 0.98 & 94.17 $\pm$ 0.22 & 43.09 $\pm$ 0.26 \\
& Mahalanobis & 51.82 $\pm$ 3.31 & 53.79 $\pm$ 0.75 & 59.52 $\pm$ 1.47 & 58.72 $\pm$ 1.25 & 82.49 $\pm$ 0.47 & \colorbox{lightgray}{68.98 $\pm$ 1.21} \\
& GNNSafe & 50.42 $\pm$ 0.86 & 35.88 $\pm$ 0.67 & 35.30 $\pm$ 0.34 & 27.35 $\pm$ 0.27 & \underline{94.82 $\pm$ 0.98} & 61.99 $\pm$ 0.96 \\
\cmidrule(lr){2-8}
& Classical ensemble (AU) & 74.00 $\pm$ 0.96 & 58.32 $\pm$ 1.19 & 58.20 $\pm$ 0.93 & 48.23 $\pm$ 0.93 & \textbf{95.30 $\pm$ 1.15} & 58.84 $\pm$ 1.46 \\
& Classical ensemble (EU) & 30.22 $\pm$ 1.23 & 59.13 $\pm$ 0.86 & \colorbox{lightgray}{65.45 $\pm$ 1.07} & 60.35 $\pm$ 0.86 & 94.81 $\pm$ 1.47 & \textbf{72.09 $\pm$ 0.86} \\
\cmidrule(lr){2-8}
& KNNLJ & 70.06 $\pm$ 0.50 & \colorbox{lightgray}{71.06 $\pm$ 0.50} & 45.35 $\pm$ 0.70 & 46.74 $\pm$ 1.09 & 32.37$\pm$ 1.07 & 36.55 $\pm$ 0.83 \\
\cmidrule(lr){2-8}
& Credal ensemble (AU) & \colorbox{lightgray}{74.98 $\pm$ 1.23} & 59.03 $\pm$ 0.86 & 58.05 $\pm$ 0.50 & 47.41 $\pm$ 0.83 & 93.85 $\pm$ 1.46 & 61.16 $\pm$ 0.50 \\
& Credal ensemble (EU) & 29.54 $\pm$ 0.70 & 53.66 $\pm$ 1.07 & 50.97 $\pm$ 0.70 & \colorbox{lightgray}{64.64 $\pm$ 1.09} & 93.72 $\pm$ 0.70 & 57.72 $\pm$ 0.50 \\
\cmidrule(lr){2-8}
& Credal final (AU) & \underline{79.29 $\pm$ 0.30} & \underline{75.02 $\pm$ 6.20} & 64.65 $\pm$ 0.20 & 68.92 $\pm$ 2.74 & 74.80 $\pm$ 6.97 & \underline{70.57 $\pm$ 1.47} \\
& Credal final (EU) & 67.27 $\pm$ 2.32 & 65.85 $\pm$ 0.57 & \underline{65.66 $\pm$ 1.27} & \underline{66.73 $\pm$ 0.23} & 50.74 $\pm$ 1.36 & 69.35 $\pm$ 2.24 \\
\cmidrule(lr){2-8}
& CredalLJ (AU) & 72.37 $\pm$ 7.16 & 73.89 $\pm$ 0.47 & \textbf{65.77 $\pm$ 0.21} & \textbf{70.78 $\pm$ 7.34} & 86.58 $\pm$ 1.43 & 66.35 $\pm$ 0.11 \\
& CredalLJ (EU) & \textbf{77.67 $\pm$ 1.32} & \textbf{77.04 $\pm$ 8.25} & 63.79 $\pm$ 0.08 & 60.06 $\pm$ 0.47 & 70.88 $\pm$ 1.92 & 67.31 $\pm$ 0.69 \\
\bottomrule
\end{tabular}%
}
\end{table}

Notably, the performance degradation on ID classification appears correlated with the number of classes. The degradation is most pronounced on datasets with a high number of classes, such as \texttt{Reddit2} (41 classes) and \texttt{Coauthor} (15 classes), while the performance is much more competitive on the 5-class datasets like \texttt{ArXiv}. This suggests that deriving a single, accurate point prediction from a credal set becomes inherently more challenging as the dimensionality of the target space grows. While methods exist to address this, such as using an Intersection Probability to map a credal set to a single probability \citep{wang2024credalwrapper}, or employing Probability Interval Dimension Reduction (PIDR) to manage the computational complexity in many-class settings \citep{wang2024credaldeep}, our focus in this work remains on the quality of the uncertainty representation itself rather than on optimizing the point-prediction decision rule.

The OOD detection results with standard deviation are presented in Table \ref{tab:ood_comparison_full}.

\section{Model selection}
\label{sec:model_selection}

\begin{table}[h]
\caption{Hyperparameter search spaces for the proposed models. $U(a, b)$ denotes a uniform distribution between $a$ and $b$.}
\label{tab:hyperparameter_tuning}
\centering
\resizebox{\textwidth}{!}{%
\begin{tabular}{lcccc}
\toprule
\textbf{Hyperparameter} & \textbf{Vanilla GNN} & \textbf{Classical \& Credal Ensemble} & \textbf{Credal Final \&  LJ} & \textbf{KNNLJ} \\
\midrule
Learning Rate (lr) & $U(10^{-5}, 10^{-1})$ & --- & $U(10^{-5}, 10^{-1})$ & --- \\
Hidden Channels & $\{64, 128, 256\}$ & --- & $\{64, 128, 256\}$ & --- \\
Num Layers & $\{2, 3\}$ & --- & $\{2, 3\}$ & --- \\
Weight Decay & $U(10^{-7}, 10^{-1})$ & --- & $U(10^{-7}, 10^{-1})$ & --- \\
GNN Type & $\{\text{GCN, SAGE}\}$ & --- & $\{\text{GCN, SAGE}\}$ & --- \\
Ensemble Size (M) & --- & $\{2, \dots, 15\}$ & --- & --- \\
Delta ($\delta$) & --- & --- & $U(0.5, 1.0)$ & --- \\
k & --- & --- & --- & $\{5, 10, 20, 50, 100, 200\}$ \\
\bottomrule
\end{tabular}
}
\end{table}

To ensure a fair and robust comparison, we performed extensive hyperparameter tuning for all baseline and proposed models, with the search spaces detailed in Table~\ref{tab:hyperparameter_tuning}. For the {Vanilla GNN} and our {Credal} models ({Final Layer} and {Latent Joint}), we employed a Bayesian optimization strategy with 30 trials for each dataset and model combination. The Vanilla GNN was optimized to maximize the validation F1-score, reflecting its primary goal of in-distribution accuracy. In contrast, our Credal models were optimized to maximize the validation AUROC based on their epistemic uncertainty score, directly tuning them for OOD detection performance. The {Classical/Credal Ensemble} baseline was constructed following a two-stage process to ensure its strength. First, we trained a large pool of 100 Vanilla GNNs using an extended Bayesian search. From this pool, we selected the top-performing models (based on their validation F1-score) to construct ensembles of varying sizes $M$. For the KNNLJ we used the best Vanilla model to compute the embeddings. 

For the baseline models, we used the hyperparameter range in their original papers.

\section{Uncertainty Decomposition in Ensemble and Bayesian Models}
\label{sec:classical_uncertainty}

Given $M$ models sampled with Bayesian Model Averaging or trained independently in an ensemble, the uncertainty can be decomposed using Shannon entropy \citep{hullermeier2021aleatoric}. The \textit{total uncertainty} TU is the entropy of the final averaged prediction, while the \textit{aleatoric uncertainty} AU is the average entropy of the individual sample predictions:
\begin{equation}
\text{TU} := H(\boldsymbol{\tilde{p}}) = -\sum_{k=1}^{C} \tilde{p}_{k} \log_{2} \tilde{p}_{k}, \quad 
\end{equation}
\begin{equation}
\text{AU} := \tilde{H}(\mathbf{p}) = \frac{1}{M} \sum_{i=1}^{M} H(\boldsymbol{p_i}) = -\frac{1}{M} \sum_{i=1}^{M} \sum_{k=1}^{C} p_{i_k} \log_{2} p_{i_k},
\end{equation}
where $C$ is the number of classes.
The \textit{epistemic uncertainty} EU is then the difference, EU $:= \text{TU} - \text{AU}$, which quantifies the disagreement among the posterior samples \citep{hullermeier2022quantification}. 

\end{document}